\PassOptionsToPackage{table}{xcolor}
\documentclass[11pt]{article}

\usepackage[final]{acl}
\usepackage{times}
\usepackage{latexsym}
\usepackage[T1]{fontenc}
\usepackage[utf8]{inputenc}
\usepackage{microtype}
\usepackage{inconsolata}
\usepackage{graphicx}
\usepackage{amsmath,amssymb,amsfonts}
\usepackage{booktabs}
\usepackage{xcolor}
\usepackage{algorithm}
\usepackage{algorithmic}
\usepackage{multirow}
\usepackage{subcaption}
\usepackage{pifont}

\newcommand{\cmark}{\ding{51}}
\newcommand{\xmark}{\ding{55}}

\newcommand{\method}{LayerMix}

\title{The Hallucination Signal Is a Mean Shift: Why Simple Probes Suffice}

\author{
  \bfseries Jungseob Lee$^{1}$ \quad
  Jaehyung Seo$^{2}$\thanks{Corresponding authors.} \quad
  Heuiseok Lim$^{1}$\footnotemark[1] \\
  $^{1}$Korea University \quad $^{2}$Konkuk University \\
  \texttt{\{omanma1928, limhseok\}@korea.ac.kr} \\
  \texttt{seojae777@konkuk.ac.kr}
}

\begin{document}
\maketitle

\begin{abstract}
Hidden-state probes effectively detect LLM hallucinations, but the geometry of the signal remains poorly characterized, driving increasingly complex probe architectures. Across three 7B-scale models and three datasets in a paired-example paradigm, we find the signal overwhelmingly dominated by a single mean-shift component, and removing this direction collapses detection to chance. Shrinkage linear discriminant analysis closes about 73\% of the gap between 1D and full-dimensional classifiers, so apparent architectural complexity largely reflects high-dimensional covariance estimation difficulty rather than exploitable non-linearity. A simple L2-regularized logistic regression (0.952 AUROC) bounds or outperforms twelve controlled architectural alternatives, and our multi-layer aggregation exceeds CLAP cross-layer attention probing under matched paradigm. Because the signal spans a contiguous layer band, \method{} aggregates it to match oracle-layer performance without oracle access. Our claims characterize the geometry within the controlled paired-example paradigm.
Our code is available at \url{https://github.com/js-lee-AI/LayerMix}.
\end{abstract}

\section{Introduction}
\label{sec:intro}

While hidden-state probing has emerged as an effective mechanism for detecting hallucinations in large language models (LLMs) \citep{burger2024truth, hernandez2023linearity, han2025simple, azaria2023internal}, recent work has pursued increasingly complex architectures to isolate the truthfulness signal \citep{luo2026two, burger2024truth, bang2025hallulens}.
Methods such as ICR Probe \citep{zhang2025icr}, TSV \citep{park2025steer}, and HaloScope \citep{du2024haloscope} deploy intricate cross-layer tracking, optimal-transport pseudo-labeling, and attention-gated subspace projections.
Yet, a fundamental question remains under-explored: \emph{what is the exact geometry of the hallucination signal, and does it necessitate this architectural complexity?}

To answer this, we conduct a systematic geometric analysis of the hallucination signal. To rigorously isolate the underlying representational structure from generation-induced distribution shifts~\citep{lee2026offpolicy}, we operate within the standard paired-example paradigm across three 7B-scale models and three datasets.
Although prior work established that truth is encoded linearly \citep{marks2023geometry,bao2025probing}, the quantitative boundaries of this linearity for hallucination detection remain unmapped.
Our analysis reveals a remarkably simple underlying structure: the detection signal is overwhelmingly dominated by a single mean-shift component.
By disentangling covariance estimation difficulty from genuine non-linear structure, we demonstrate that a simple L2-regularized logistic regression (L2-LR) probe matches or outperforms 12 heavily engineered alternatives. This suggests that complex architectures often overfit to estimation noise rather than exploiting hidden structural complexities.

Crucially, we observe that this simple mean-shift signal is not isolated to a single oracle layer, but rather distributed across a model-specific, contiguous layer band~\citep{meng2022locating, patrawala2025llm, huang2025survey}. Motivated directly by this geometric insight, we operationalize our findings into \method{}, a computationally lightweight, cross-validation-based multi-layer aggregation approach. \method{} is not presented as a complex algorithmic innovation, but rather as a principled consequence of our analysis: it exploits the distributed nature of the signal to match oracle-layer performance without requiring oracle access, offering a robust, highly effective drop-in baseline for future geometric research.

Our contributions are as follows:
\begin{itemize}
    \item \textbf{Quantitative Geometric Decomposition:} We provide the first quantitative characterization of the hallucination signal's geometry, demonstrating that a dominant mean-shift accounts for the vast majority of the signal. We show that shrinkage linear discriminant analysis (LDA) closes about 73\% of the Fisher LDA gap, indicating that apparent structural complexity largely stems from high-dimensional covariance estimation difficulty rather than exploitable non-linearities.
    
    \item \textbf{Controlled Geometric Ablation:} To rigorously validate our geometric framework, we evaluate L2-LR against 12 controlled architectural alternatives designed to isolate specific geometric properties, alongside complex subspace methods. As predicted by our decomposition, L2-LR strictly upper-bounds their performance, achieving 0.952 AUROC and demonstrating that mean-targeting methods consistently dominate variance-targeting ones.
    
    \item \textbf{\method{}:} We introduce a principled, oracle-free multi-layer aggregation strategy. \method{} seamlessly identifies and aggregates the informative layer band, matching oracle performance (0.954 AUROC) while remaining exceptionally computationally efficient (about 35 seconds overhead).
\end{itemize}

The paired-example paradigm is the controlled experimental setup we use to isolate the intrinsic representational geometry from generation-induced distribution shifts, the methodological control that makes the geometric measurement possible. The conclusions reported throughout (mean-shift dominance, decision-boundary linearity, multi-layer aggregation sufficiency) are paradigm-bounded statements about the controlled measurement setting, not universality claims across all detection scenarios.

\section{Related Work}
\label{sec:related}

\subsection{Detection Methods and Evaluation}

\paragraph{Training-free detection.}
Methods requiring no labeled data span logit contrasting~\citep{chuang2023dola}, covariance eigenvalues~\citep{chen2024inside}, semantic entropy~\citep{kuhn2023semantic}, and sampling consistency~\citep{manakul2023selfcheckgpt}, with LLM-Polygraph~\citep{fadeeva2023lm} providing a comprehensive benchmark.
The limited accuracy of these approaches motivates our focus on supervised probing.

\paragraph{Probing-based detection.}
Following \citet{azaria2023internal}'s foundational logistic regression probe, recent methods have introduced increasingly complex architectures, including attention-gated SVD~\citep{du2024haloscope}, cross-layer tracking~\citep{zhang2025icr}, and optimal-transport pseudo-labeling~\citep{park2025steer}.
Crucially, these single-layer methods lack a geometric characterization of the signal they target.
Our analysis fills this gap, demonstrating that much of this architectural complexity is unwarranted when using L2-regularized logistic regression.

\paragraph{Evaluation paradigms.}
SEP~\citep{kossen2024semantic} trains probes to predict semantic entropy from multiple sampled generations.
This open-ended QA paradigm differs fundamentally from the paired-example setting evaluated here (and in SAPLMA, ICR Probe, and HaloScope), precluding direct head-to-head comparison. We discuss SEP's cross-domain implications in \S\ref{sec:discussion}.

\subsection{Geometry and Layer Choice}

\paragraph{Geometry of LLM representations.}
Prior work has identified linear structures for truthfulness~\citep{marks2023geometry,bao2025probing,burns2022discovering} and world representations~\citep{li2022emergent}.
Representation engineering~\citep{zou2023representation,lee2026skindeep,lee2026refusal} shows that projecting along such directions steers model behavior, a principle our intervention experiment (\S\ref{sec:discussion}) applies to hallucinations.
While HARP~\citep{hu2025harp} derives hallucination subspaces from unembedding weights, we extend this trajectory by decomposing the detection signal into mean-shift and residual components, quantifying the Fisher LDA gap, and linking geometry to probe performance.

\paragraph{Layer selection.}
While probing accuracy heavily depends on the chosen hidden layer~\citep{azaria2023internal,han2025simple,belinkov2022probing}, the exact cost of relying on non-oracle layer heuristics remains unquantified for hallucination detection.
We systematically measure this degradation and demonstrate that cross-validation-based aggregation eliminates it.

\subsection{Concurrent Work}

\paragraph{Cross-layer and subspace probing.}
The closest concurrent work to \method{} is CLAP \citep{clap2024}, which learns input-dependent cross-layer attention, whereas \method{} selects a layer band by cross-validation and uniformly averages independent per-layer probes.
We run CLAP from its official implementation on our cached multi-layer hidden states, so the contrast isolates the combination rule, not the features.
\citet{burger2024truth} argue instead that hallucination and truth occupy a low-dimensional subspace, which we test with partial least squares discriminant analysis (PLS-DA) at $k{=}2$ against full-dimensional probes.
Our shrinkage-LDA-gap analysis predicts that the residual beyond the mean shift needs high-dimensional shrinkage, not two-dimensional truncation, and \S\ref{sec:results} reports both comparisons under matched paradigm.

\paragraph{Findings in other paradigms.}
\citet{wei2025drift} report SOTA on free-form generation with cross-dataset generalization, which is not in tension with our orthogonal-mean-shift finding, because DRIFT operates in the dynamic-generation paradigm, orthogonal to our paired-example setting, where reweighting across directions can succeed even when raw paired-example $\hat{\boldsymbol{\delta}}$ are orthogonal across datasets.
\citet{liang2025mlpprobes} report MLP probes outperforming linear probes in token-level free-form detection, which we likewise read as paradigm-dependence rather than a contradiction.

\paragraph{Generalization and controls.}
\citet{orgad2025reps} document that representation-based detectors do not generalize zero-shot across datasets.
Our within-vs-between cosine analysis (Appendix~\ref{app:within_vs_between}) suggests this reflects hallucination-type-specific geometry rather than pure dataset artifacts.
Our random-label diagnostic follows probing-classifier control tasks~\citep{hewitt2019designing}, formalizing selectivity for hallucination detection.

\section{Geometric Analysis}
\label{sec:geometry}

We characterize the hallucination signal's geometry to motivate \method{}.
Given an LLM with hidden dimension $d$ and labeled examples $\{(x_i, y_i)\}$ where $y_i \in \{0,1\}$ indicates factual or hallucinated, let $\boldsymbol{\mu}_0, \boldsymbol{\mu}_1$ be class centroids and $\boldsymbol{\delta} = \boldsymbol{\mu}_1 - \boldsymbol{\mu}_0$ the mean-shift direction.

\subsection{The Signal is a Mean Shift}

Table~\ref{tab:meanshift} provides direct evidence from decomposition experiments averaged over 3 models $\times$ 3 datasets. Removing $\boldsymbol{\delta}$ collapses detection to chance (0.499), confirming it is \emph{necessary}. Crucially, $\boldsymbol{\delta}$ is estimated within each training fold of the CV protocol, preventing in-sample estimation artifacts. 

The 1D projection onto $\boldsymbol{\delta}$ achieves 0.834 AUROC, recovering 78\% of the above-chance performance of unregularized LR. The gap between this and the operational L2-LR baseline (0.952) highlights the regularization benefit: L2 shrinkage suppresses noise in the roughly 4,000 residual dimensions, extracting signal that unregularized LR cannot reliably exploit. Note that PLS-DA component~1 aligns with $\boldsymbol{\delta}$ by mathematical construction in the binary case~\citep{barker2003partial}, confirming that PLS-DA inherently targets the mean shift.

\begin{figure}[t]
    \centering
    \includegraphics[width=0.9\columnwidth]{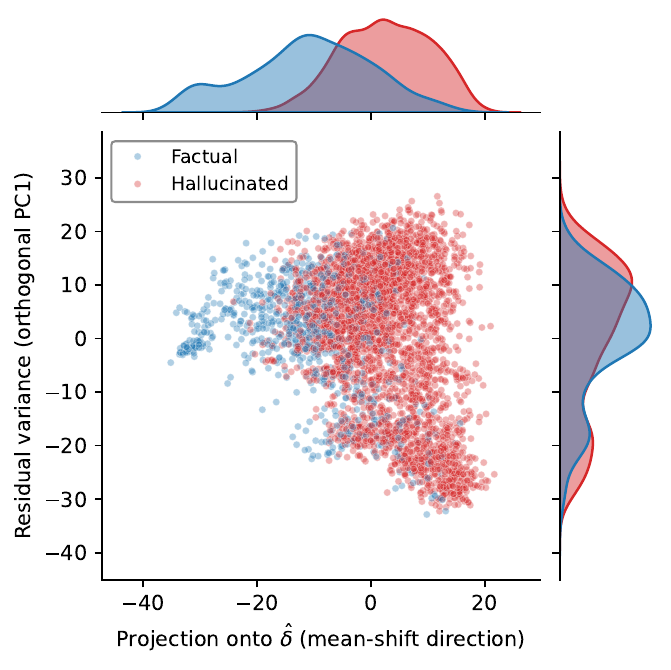}
    \caption{2D projection of factual and hallucinated hidden states from Qwen2.5-7B (Layer 18) on TruthfulQA.}
    \label{fig:geometry_projection}
\end{figure}

Cohen's $d$ along $\boldsymbol{\delta}$ ranges 1.2--1.6 despite accounting for ${<}3\%$ of total variance. Figure~\ref{fig:geometry_projection} visually corroborates this: the clear macro-separation along the mean-shift axis contrasts sharply with the completely overlapping distributions in the residual orthogonal subspace, explaining why linear probes successfully recover the vast majority of the signal.

\begin{table}[t]
\centering
\small
\begin{tabular}{lc}
\toprule
\textbf{Condition} & \textbf{AUROC} \\
\midrule
Unreg.\ LR (all dims) & .928 \\
Mean-shift only (1D) & .834 \\
Mean-shift + 4 residual (5D) & .891 \\
\midrule
Without mean-shift & .499 \\
Mean-centered (oracle) & .503 \\
\bottomrule
\end{tabular}
\caption{Mean-shift decomposition (avg.\ 9 conditions). ``Unreg.\ LR'' denotes unregularized logistic regression, used to enable clean decomposition into mean-shift and residual components without regularization-induced shrinkage.}
\label{tab:meanshift}
\end{table}

\subsection{Mean Shift is Necessary but Insufficient}

The mean-shift direction is essential but captures only part of the signal.

\paragraph{Fisher LDA gap.}
In the binary case, Fisher LDA projects onto $\boldsymbol{w}_F = \Sigma_W^{-1}\boldsymbol{\delta}$, which reduces to $\boldsymbol{\delta}$ only when the within-class covariance $\Sigma_W$ is proportional to the identity. In high dimensions ($d \approx 4{,}000$), $\Sigma_W$ is far from spherical, so $\boldsymbol{w}_F$ and $\boldsymbol{\delta}$ differ. We report projection onto the raw mean-shift direction $\boldsymbol{\delta}$ to provide a clean lower bound on the 1D signal without requiring covariance inversion. 

Table~\ref{tab:fisher_gap} shows that this optimal 1D classifier underperforms L2-regularized LR by 0.09--0.16 AUROC. This consistent gap confirms the presence of discriminative structure beyond $\boldsymbol{\delta}$ in every condition. To disentangle covariance estimation difficulty from genuine distributional structure, we apply shrinkage LDA~\citep{ledoit2004well}, which achieves 0.920 mean AUROC, closing at least 73\% of the gap. The residual 0.032 gap may reflect non-Gaussian structure or L2-regularization's implicit feature selection. Applying ZCA whitening ($\Sigma^{-1/2}$) before L2-LR degraded performance (0.790 vs.\ 0.952 raw), as expected when amplifying low-variance noise in finite high-dimensional samples.

\paragraph{Distributional evidence.}
D'Agostino--Pearson tests reject normality ($p < 10^{-4}$) for both classes in all 9 conditions, with skewness $|\gamma_1| \leq 0.55$ and excess kurtosis $-1.1$ to $+0.2$. This is consistent with the residual gap reflecting distributional structure beyond what LDA can capture.

\begin{table}[t]
\centering
\small
\begin{tabular}{lccc}
\toprule
\textbf{Condition} & \textbf{Fisher} & \textbf{L2-LR} & \textbf{Gap} \\
\midrule
Llama--TQA & .815 & .937 & +.122 \\
Llama--HE & .872 & .963 & +.090 \\
Llama--FD & .823 & .951 & +.127 \\
Mistral--TQA & .824 & .940 & +.116 \\
Mistral--HE & .873 & .960 & +.088 \\
Mistral--FD & .827 & .955 & +.128 \\
Qwen--TQA & .815 & .943 & +.128 \\
Qwen--HE & .852 & .961 & +.109 \\
Qwen--FD & .802 & .957 & +.156 \\
\midrule
\textbf{Mean} & \textbf{.834} & \textbf{.952} & \textbf{+.118} \\
\bottomrule
\end{tabular}
\caption{Raw mean-shift projection (1D, onto $\boldsymbol{\delta}$) vs.\ L2-regularized LR on the optimal layer across all 9 conditions.}
\label{tab:fisher_gap}
\end{table}

\subsection{Signal Distributes Across Layers}
\label{sec:cross_layer}

\begin{figure*}[t]
\centering
\includegraphics[width=0.9\textwidth]{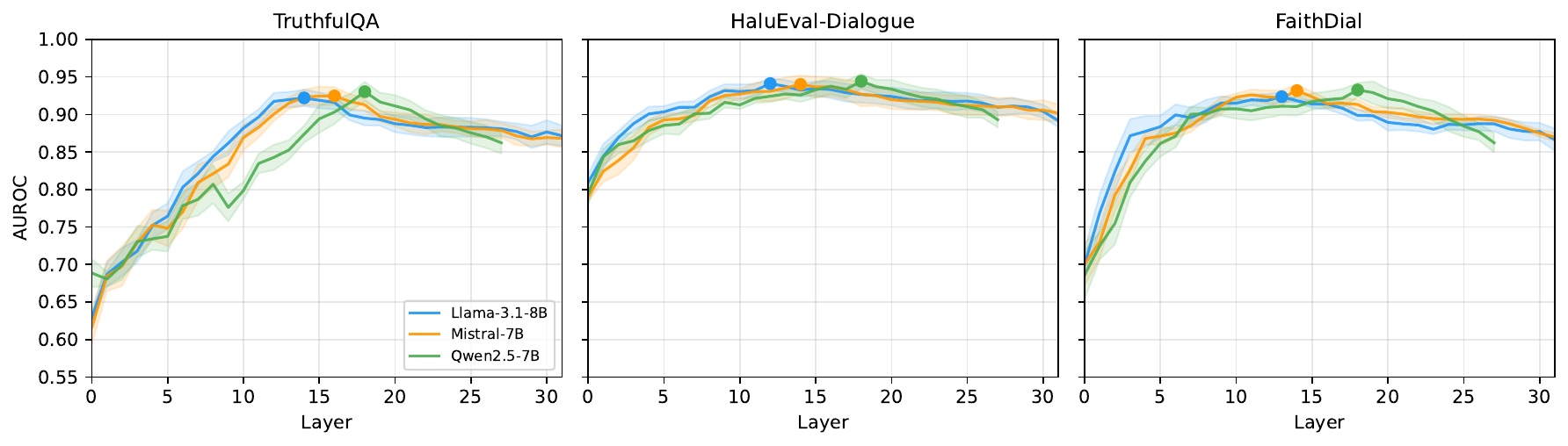}
\caption{Layer-wise AUROC across 3 models and 3 datasets. Shaded bands show ${\pm}1$ std across CV folds. Dots mark the optimal layer per condition.}
\label{fig:layer_auroc}
\end{figure*}

Beyond characterizing the signal within a single layer, we examine how it distributes \emph{across} layers to motivate multi-layer probing:

\paragraph{Optimal layer varies.}
The best single layer differs by model: L14 for Llama~\citep{touvron2023llama2openfoundation, grattafiori2024llama}, L16 for Mistral~\citep{jiang2023mistral7b}, L18 for Qwen~\citep{qwen2025qwen25technicalreport} (Figure~\ref{fig:layer_auroc}; Table~\ref{tab:layer_sensitivity}). This variation means a fixed-layer heuristic cannot be optimal across conditions. As Figure~\ref{fig:layer_auroc} illustrates, the hallucination signal instead concentrates in a model-specific contiguous band.

\paragraph{Adjacent layers carry signal.}
Cross-validation scores for layers near the optimum remain high. For example, on Qwen--TQA (optimal L18), CV selects layers $\{17, 18, 19, 20, 21\}$, each scoring $\geq$0.93.

\paragraph{Signal distributes across neurons.}
L1-regularized feature selection followed by L2-LR (Appendix~\ref{app:sparse_probing}) shows that 200 neurons (about 5\% of $d$) recover 98.3\% of full AUROC, while 100 random neurons achieve only 0.840. The top-20 neurons share only 4 across CV folds, confirming a distributed signal not localized to fixed "hallucination neurons."

\paragraph{Combining layers improves detection.}
Averaging predictions from CV-selected layers consistently outperforms any single layer (by +0.0004 to +0.0086 AUROC across all 9 conditions), indicating systematic complementarity. 

These findings establish that the hallucination signal distributes across both layers and neurons, motivating multi-layer aggregation of full-dimensional probes. The geometry also correctly predicts the performance ordering of subspace methods (Table~\ref{tab:method_taxonomy} in Appendix~\ref{app:method_taxonomy}): methods targeting the mean shift (PLS-DA: 0.926) consistently outperform variance-targeting ones (SVD: 0.795).

\section{Proposed Method: \method{}}
\label{sec:method}

Our geometric analysis reveals that the hallucination signal distributes across a contiguous band of layers (\S\ref{sec:cross_layer}), with the optimal layer varying by model and dataset.
This motivates \method{}: a principled, computationally lightweight approach that identifies the informative layer band via cross-validation (CV), trains independent probes on the top-$K$ layers, and averages their predictions.
The method operates in three stages.

\paragraph{Stage 1: Layer scoring.}
For each layer $\ell \in \{1, \ldots, L\}$, we extract hidden states $\mathbf{H}^{(\ell)} \in \mathbb{R}^{N \times d}$ and evaluate an L2-regularized logistic regression probe via stratified $m$-fold CV ($m{=}5$).
The score $s_\ell$ is the mean AUROC across folds.
This step replaces oracle selection: rather than requiring a held-out set to identify the best layer post hoc, CV scores provide a data-driven ranking using only training data.

\paragraph{Stage 2: Layer selection.}
We select the top-$K$ layers: $\mathcal{L}^* = \text{top-}K(\{s_\ell\}_{\ell=1}^L)$.
We use $K{=}5$ as the default, finding it robust across conditions (Appendix~\ref{app:layermix_ablation}).
In practice, the selected layers naturally form a contiguous block around the model's most informative region (e.g., layers 16--20 for Qwen2.5-7B), as adjacent layers share representational structure while providing marginal complementary signal.

\paragraph{Stage 3: Aggregation.}
For each selected layer $\ell \in \mathcal{L}^*$, we train an L2-regularized LR probe $f_\ell$ on the full training set.
The final prediction for a test input $x$ is:
\begin{equation}
    \hat{p}(x) = \frac{1}{K} \sum_{\ell \in \mathcal{L}^*} f_\ell(\mathbf{h}^{(\ell)}(x))
\end{equation}

The simplicity of this formulation is deliberate: each per-layer probe operates on the full $d$-dimensional hidden state with strong L2 regularization ($C{=}0.001$), and uniform averaging provides implicit variance reduction.
No learned aggregation weights are needed, as the selected layers are high-quality by construction.

\begin{algorithm}[t]
\caption{\method{}}
\label{alg:layermix}
\small
\begin{algorithmic}[1]
\REQUIRE LLM $\mathcal{M}$, labeled data $\mathcal{D} = \{(x_i, y_i)\}$, $K{=}5$
\ENSURE Detector $f: x \to [0, 1]$
\STATE \textbf{// Stage 1: Layer Scoring (replaces oracle selection)}
\FOR{$\ell = 1, \ldots, L$}
    \STATE $\mathbf{H}^{(\ell)} \leftarrow$ hidden states from layer $\ell$
    \STATE $s_\ell \leftarrow$ CV-AUROC$(\text{LR}(\mathbf{H}^{(\ell)}, \mathbf{y}; C{=}0.001))$
\ENDFOR
\STATE \textbf{// Stage 2: Layer Selection}
\STATE $\mathcal{L}^* \leftarrow \text{top-}K(\{s_\ell\}_{\ell=1}^L)$
\STATE \textbf{// Stage 3: Per-Layer Probes + Aggregation}
\FOR{$\ell \in \mathcal{L}^*$}
    \STATE Train $f_\ell \leftarrow \text{LR}(\mathbf{H}^{(\ell)}, \mathbf{y};\ C{=}0.001)$
\ENDFOR
\STATE $f(x) \leftarrow \frac{1}{K}\sum_{\ell \in \mathcal{L}^*} f_\ell(\mathbf{h}^{(\ell)}(x))$
\end{algorithmic}
\end{algorithm}

\paragraph{Design choices.}
(1)~\emph{CV-based selection} identifies layers by their actual classification performance, unlike heuristics (last, middle) or statistics such as Cohen's $d$ ranking, which we show fails systematically (Appendix~\ref{app:layermix_ablation}).
(2)~\emph{Full-dimensional probes} avoid the information loss of dimensionality reduction; our geometric analysis confirms L2-regularized LR outperforms subspace methods on single layers.
(3)~\emph{Prediction averaging} prevents the dimensionality from scaling with $K$ (unlike feature concatenation) and provides ensemble-style variance reduction.

\paragraph{Computational cost.}
\method{} requires extracting hidden states from all $L$ layers during the scoring phase (a single forward pass) and training $K$ logistic regression probes.
At inference, hidden states from $K{=}5$ layers are extracted during the standard forward pass (no additional passes required).
The total wall-clock overhead is minimal: scoring all layers takes about 30 seconds on a single GPU for $N{=}5{,}000$ examples, and the $K$ probe trainings add only about 5 seconds. We note that an alternative held-out layer sweep is \emph{not cheaper}: it requires (a) sacrificing held-out data and (b) the same per-layer LR fits as our CV scoring. CV-based selection therefore matches the held-out sweep in compute while preserving data efficiency.

\section{Experimental Setup}
\label{sec:setup}

\paragraph{Models \& Datasets.}
We evaluate three 7B-scale base LLMs: \textbf{Llama-3.1-8B}~\citep{grattafiori2024llama}, \textbf{Mistral-7B}~\citep{jiang2023mistral7b}, and \textbf{Qwen2.5-7B}~\citep{qwen2025qwen25technicalreport}, scaling up to 70B in \S\ref{sec:instruct}.
We use three standard benchmarks: \textbf{TruthfulQA} (TQA;~\citealp{lin2022truthfulqa}, $N{=}4{,}135$), \textbf{HaluEval-Dialogue} (HE;~\citealp{li2023halueval}, $N{=}20{,}000$), and \textbf{FaithDial} (FD;~\citealp{dziri2022faithdial}, $N{=}5{,}848$).

\paragraph{Baselines.}
We compare against a broad spectrum of methods: (1)~\textbf{Training-free}: Perplexity~\citep{ren2022out}, Entropy, Self-eval P(True)~\citep{kadavath2022language}, Verbalize~\citep{lin2022teaching}, LLM-Check~\citep{sriramanan2024llm}, DoLa~\citep{chuang2023dola}, INSIDE~\citep{chen2024inside}; (2)~\textbf{Unsup./Semi-sup.}: HaloScope~\citep{du2024haloscope}, CCS~\citep{burns2022discovering}, TSV~\citep{park2025steer}; (3)~\textbf{Cross-layer}: ICR Probe~\citep{zhang2025icr}; (4)~\textbf{Subspace}: SVD, MeanDiff~\citep{zou2023representation}, PLS-DA, gcPCA~\citep{abid2018exploring}; (5)~\textbf{Supervised}: SAPLMA~\citep{azaria2023internal}, Oracle-layer LR, and our \textbf{\method{}}.

\paragraph{Protocol.}
We employ 5-fold stratified CV. To isolate the representational geometry from generation-induced distribution shifts, we evaluate within the standard paired-example paradigm. Hidden states are extracted at the last token position of each input sequence. \method{} scoring uses nested 5-fold CV within training folds, ensuring strictly held-out selection. All methods use StandardScaler on training data. \textbf{Length confound check.} Length-only AUROC (word-length of the answer/response token sequence) remains far below probe AUROC across all three datasets: TQA 0.546, HE 0.610, FD 0.511. Probes consistently exceed length-only AUROC by $\geq 0.34$, ruling out length shortcuts.

\section{Results}
\label{sec:results}

\subsection{Main Comparison}
\label{sec:results_main}
\label{subsec:main_comparison}
Table~\ref{tab:main} presents detection AUROC averaged across three models. \method{} achieves 0.954 mean AUROC across 9 conditions, fully matching the single-layer oracle (0.952). Crucially, \method{} captures this distributed mean-shift signal perfectly without requiring post-hoc oracle layer access, eliminating a major bottleneck for practical geometric analysis and deployment.

The best zero-overhead heuristic, middle-layer LR, achieves 0.941. The 0.013 gap closed by \method{} is statistically significant ($p < 0.001$) in 8/9 conditions. Practically, at a fixed recall of 0.90, this AUROC improvement translates to a 15--20\% relative reduction in false positives. For practitioners without labeled data, the middle-layer heuristic provides a strong zero-cost alternative; however, \method{}'s principled aggregation adds only modest cost (a single forward pass and about 35 seconds for training) when maximum fidelity to the geometric signal is required.

All-layer averaging (0.944) underperforms \method{}, confirming that structurally informed selection adds value beyond naive ensembling. Against the two concurrent designs discussed in \S\ref{sec:related}, \method{} exceeds CLAP in 9/9 conditions (0.954 vs.\ 0.928 mean AUROC), and the two-component PLS-DA subspace probe reaches 0.910, under-performing full-dimensional probes by 0.044. Furthermore, among single-layer subspace methods, PLS-DA (0.941) clearly outperforms SVD (0.929). This strongly corroborates our theoretical geometric decomposition (\S\ref{sec:geometry}), demonstrating that methods explicitly targeting the mean-shift vector strictly dominate those capturing generic variance.

\paragraph{Standardized evaluation context as an in-vitro control.}
\label{sec:baseline_disc}
As shown in Table~\ref{tab:main}, several baselines underperform their originally published numbers. All methods there are evaluated on identical pre-extracted hidden states under our paired-example paradigm, so the differences reflect each method's interaction with the intrinsic representational geometry rather than its degree of paradigm-specific tuning. Methods originally designed for dynamic generation (HaloScope, ICR Probe, TSV, CLAP) are therefore evaluated outside their native deployment regime, and their published numbers under that regime are not in question. 

To ensure a strictly fair comparison under these equated geometric conditions, we re-evaluated these baselines on instruct models with matched L2 regularization ($C{=}0.001$; Table~\ref{tab:fairness}). Proper regularization significantly improves SEP (0.889 $\to$ 0.921) and the oracle LR (0.935 $\to$ 0.957). Nevertheless, \method{} (0.960) retains a clear advantage over SEP and edges out the matched oracle LR, confirming its benefits derive from capturing the distributed multi-layer structure rather than favorable hyperparameter tuning.

\begin{table}[t]
\centering
\footnotesize
\renewcommand{\arraystretch}{1.1}
\setlength{\tabcolsep}{4pt}
\resizebox{\columnwidth}{!}{%
\begin{tabular}{ll cccc}
\toprule
\textbf{Method} & \textbf{Type} & \textbf{TQA} & \textbf{HE} & \textbf{FD} & \textbf{Mean} \\
\midrule
\multicolumn{6}{c}{\textbf{\textit{Training-free baselines}}} \\
\midrule
Perplexity & unsup. & .594 & .445 & .560 & .533 \\
Entropy & unsup. & .576 & .476 & .536 & .529 \\
Self-eval P(True) & unsup. & .621 & .812 & .548 & .660 \\
Verbalize & unsup. & .618 & .768 & .541 & .642 \\
LLM-Check & unsup. & .500 & .500 & .500 & .500 \\
DoLa & unsup. & .607 & .553 & .579 & .580 \\
INSIDE & unsup. & .559 & .594 & .542 & .565 \\
\midrule
\multicolumn{6}{c}{\textbf{\textit{Unsupervised / semi-supervised probes}}} \\
\midrule
HaloScope$^\dagger$ & unsup. & .576 & .658 & .584 & .606 \\
CCS & unsup. & .564 & .519 & .528 & .537 \\
TSV$^\dagger$ & semi-sup. & .812 & .677$^*$ & .642$^*$ & .710 \\
ICR Probe & sup. & .567 & .628 & .540 & .578 \\
\midrule
\multicolumn{6}{c}{\textbf{\textit{Subspace probes (single layer, oracle)}}} \\
\midrule
SVD+LR ($k{=}100$) & sup. & .925 & .940 & .921 & .929 \\
PLS-DA+LR ($k{=}2$) & sup. & .908 & .921 & .900 & .910 \\
PLS-DA+LR ($k{=}5$) & sup. & .926 & .948 & .948 & .941 \\
\midrule
\multicolumn{6}{c}{\textbf{\textit{Full-dim probes (single layer)}}} \\
\midrule
SAPLMA (MLP) & sup. & .908 & .930 & .918 & .919 \\
Last-layer LR & sup. & .895 & .936 & .917 & .918 \\
Middle-layer LR & sup. & .924 & .954 & .946 & .941 \\
Oracle LR & sup. & \underline{.940} & \underline{.961} & \underline{.954} & \underline{.952} \\
\midrule
\multicolumn{6}{c}{\textbf{\textit{Multi-layer aggregation}}} \\
\midrule
All-layer avg & sup. & .916 & .957 & .958 & .944 \\
CLAP$^a$ & sup. & .896 & .953 & .935 & .928 \\
\rowcolor{gray!10}
\textbf{\method{} (Ours)} & sup. & \textbf{.942} & \textbf{.962} & \textbf{.957} & \textbf{.954} \\
\bottomrule
\end{tabular}
}
\caption{Hallucination detection AUROC averaged over three 7B-scale base models. Bold marks \method{} and underline marks the single-layer oracle. $\dagger$Official implementation. $^*$Adapted for dataset-specific pseudo-labeling. $^a$\citet{clap2024} cross-layer attention probe, trained from the official implementation on the same cached multi-layer hidden states under 3-fold cross-validation.}
\label{tab:main}
\end{table}

\begin{table}[t]
\centering
\small
\begin{tabular}{lcccc}
\toprule
\textbf{Method} & \textbf{TQA} & \textbf{HE} & \textbf{FD} & \textbf{Avg} \\
\midrule
\multicolumn{5}{l}{\textit{Last-layer probes}} \\
SEP ($C{=}1.0$, current) & .888 & .901 & .879 & .889 \\
SEP ($C{=}0.001$) & .915 & .941 & .906 & .921 \\
MLP PCA(64) (current) & .895 & .912 & .843 & .883 \\
MLP no-PCA & .913 & .923 & .889 & .908 \\
\midrule
\multicolumn{5}{l}{\textit{Oracle-layer probes}} \\
LR oracle ($C{=}1.0$) & .932 & .935 & .937 & .935 \\
LR oracle ($C{=}0.001$) & .951 & .965 & .954 & .957 \\
MLP oracle PCA(64) & .942 & .948 & .926 & .938 \\
MLP oracle no-PCA & .949 & .956 & .947 & .950 \\
\midrule
\textbf{\method{}} & \textbf{.953} & \textbf{.966} & \textbf{.957} & \textbf{.960} \\
\bottomrule
\end{tabular}
\caption{Fairness ablation on instruct models (3-model average AUROC). Performance is compared across methods under matched L2 regularization ($C{=}0.001$).}
\label{tab:fairness}
\end{table}

\subsection{Ablation Studies}
\label{sec:ablations}

\paragraph{Heuristic baselines.}
Every fixed-rule layer heuristic underperforms \method{} (Table~\ref{tab:layermix_ablation}c in Appendix~\ref{app:layermix_ablation}). Notably, CV-based selection is the only strategy that consistently matches or outperforms the single-layer oracle, whereas metric-based rankings like Cohen's $d$ select scattered layers rather than the optimal contiguous band where the mean-shift is most pronounced.

\paragraph{Number of layers.}
Performance is highly stable across $K \in \{3, 5, 7\}$ (differences $\leq 0.002$; Table~\ref{tab:layermix_ablation}b). We adopt $K{=}5$ as the default.

\paragraph{Label efficiency.}
At $N{\leq}200$, PLS-DA $k{=}3$ outperforms full-dimensional LR (Appendix~\ref{app:fewshot}), as the low-rank constraint acts as an implicit regularizer. At $N{\geq}500$, full-dimensional LR overtakes PLS-DA.

\paragraph{Statistical significance \& Robustness.}
Paired bootstrap tests confirm \method{} vs.\ last layer reaches $p < 0.001$ in 9/9 conditions. A stricter nested CV protocol (outer 5-fold, inner 3-fold) confirms negligible optimistic bias ($\leq$0.010). Random-label controls yield chance AUROC ($0.500 \pm 0.003$).

\subsection{The Decision Boundary is Linear in the Paired-Example Setting}
\label{sec:linearity}

A controlled MLP experiment (256 units, ReLU, dropout 0.3) on the oracle layer yields an absolute difference of $|\Delta| \leq 0.002$ AUROC vs.\ L2-LR across all 9 conditions (Appendix~\ref{app:mlp_per_condition}). This demonstrates that, within the paired-example paradigm, the class-discriminative boundary is overwhelmingly linear, explaining why SAPLMA's 2-layer MLP (0.919) underperforms a properly regularized LR (0.952). The complexity of the signal does not warrant non-linear functional forms in this setting. We note that \citet{liang2025mlpprobes} report MLP probes outperforming linear probes in token-level free-form detection; on our reading this reflects a paradigm difference (paired-example sequence-level vs.\ token-level free-form) rather than a contradiction.

\subsection{Testing Geometric Hypotheses via Controlled Ablation}
\label{sec:complex_methods}

Our geometric framework posits that methods discarding raw hidden states or projecting onto generic variance-maximizing subspaces will inherently lose the mean-shift signal. Because evaluating existing SOTA architectures often conflates multiple algorithmic design choices, we rigorously tested our structural predictions by evaluating 12 controlled alternative architectures (Appendix~\ref{app:full_geometric_ablation}). These were explicitly designed to isolate and ablate specific geometric properties (e.g., cross-layer dynamics tracking, domain-adversarial transforms, discriminant subspaces). By evaluating these isolated factors within our controlled paradigm, we empirically validate the theoretical findings of Section 5: none of the 12 complex alternatives improve upon the simple L2-LR (0.952 in-domain). Their performance ordering strictly correlates with how explicitly they preserve the mean-shift component, providing evidence that complex architectures often overfit to domain-specific covariance noise rather than exploiting hidden non-linear structures.

\subsection{Instruction-Tuned Models and Scaling}
\label{sec:instruct}

\begin{figure}[t]
    \centering
    \includegraphics[width=\columnwidth]{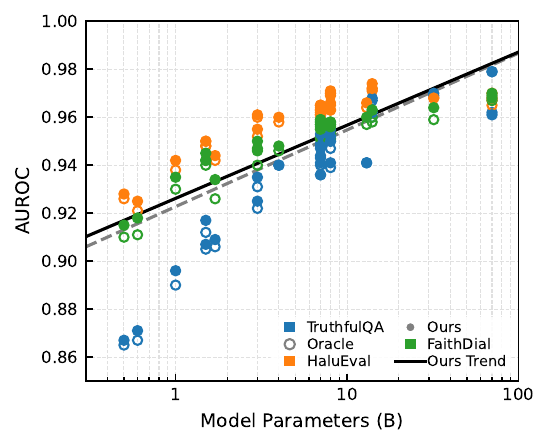}
    \caption{Scaling law of hallucination detection across 25 models (0.5B to 70B), evaluated on 75 conditions (25 models $\times$ 3 datasets). The plot compares the performance of \method{} (solid markers and trend line) against the single-layer Oracle (hollow markers and dashed trend line).}
    \label{fig:scaling_law}
\end{figure}

To verify generalization, we map the scaling trajectory across 25 models spanning 5 families (0.5B to 70B) in Figure~\ref{fig:scaling_law}.

First, \textbf{hallucination detection scales predictably with model capacity}. Across all model families, we observe a monotonic increase in AUROC as parameter count grows. Small models (${<}2$B) start in the 0.86--0.91 range, while massive models (e.g., Llama-3.1-70B) approach 0.98 AUROC. This implies that the linear mean-shift structure becomes increasingly prominent and separable as the model's representational capacity expands.

Second, \textbf{instruction tuning crystallizes the hallucination signal}. On Llama-3.1-8B-Instruct, the oracle AUROC rises to 0.959 (from 0.951 base), and \method{} achieves 0.961. This confirms that the mean-shift geometry is not merely a pretraining artifact but is preserved and even sharpened during alignment.

Third, \textbf{\method{} scales robustly}. As depicted, our method tightly tracks or exceeds the single-layer oracle across 72 of 75 evaluated conditions (96.0\%). This demonstrates that multi-layer aggregation remains highly effective from 0.5B up to the 70B scale without necessitating computationally expensive held-out layer selection.

\section{Discussion and Conclusion}
\label{sec:discussion}

\paragraph{Scope and predictions.}
Our geometric analysis within the paired-example paradigm yields a testable prediction: approaches discarding raw hidden states or projecting onto low-dimensional subspaces should underperform properly regularized LR. This holds across all 12 hypothesis-driven alternatives tested (\S\ref{sec:complex_methods}). While recent work corroborates linear truth-directions~\citep{bao2025probing,marks2023geometry}, our contribution is quantifying that the Fisher LDA gap primarily reflects covariance estimation difficulty (about 73\%) rather than exploitable residual structure.

\paragraph{Implications for method design.}
\method{} exploits the signal's distribution across a contiguous layer band, precisely where prediction averaging succeeds~\citep{breiman1996bagging} (9/9 conditions). Within this paradigm, L2-regularized LR establishes a rigorous baseline that future architectural proposals must match.

\paragraph{When to use probing-based detection.}
While probing requires white-box access and labeled data, it drastically outperforms training-free methods (0.95+ vs.\ 0.50--0.66 AUROC). For practitioners, middle-layer LR at $\lfloor L/2 \rfloor$ provides a strong zero-overhead default (0.941), while \method{} adds 0.013 AUROC at modest cost (about 35 seconds) when maximal accuracy is required.

\begin{figure}[t]
    \centering
    \includegraphics[width=0.9\columnwidth]{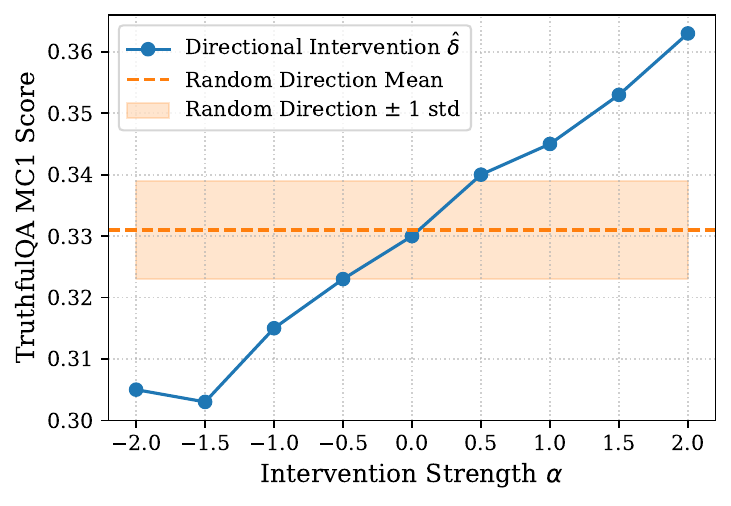}
    \caption{Intervention alpha sweep on Qwen2.5-7B (Layer 18, TruthfulQA MC1). The plot shows the effect of projecting hidden states along the mean-shift direction $\hat{\boldsymbol{\delta}}$ across varying $\alpha$ values, compared against 10 random control directions.}
    \label{fig:intervention_sweep}
\end{figure}

\paragraph{Is the mean shift an artifact?}
One might worry that the dominant mean shift reflects superficial features (e.g., length or style) rather than a genuine hallucination signal~\citep{levinstein2025still}. Five observations argue against this: (1)~length-only AUROC ($\le0.610$) is far below probe performance, and regressing out length changes AUROC by $\le0.003$; (2)~reinforcing $\hat{\boldsymbol{\delta}}$ degrades truthfulness while projecting away improves it (Figure~\ref{fig:intervention_sweep}), providing bidirectional causal evidence; (3)~random-label controls yield chance AUROC; (4)~the mean-shift dominance replicates exactly across Llama, Mistral, and Qwen; (5)~removing $\boldsymbol{\delta}$ collapses detection to chance across all 9 conditions. While confounds within the paired-example paradigm cannot be definitively excluded without free-form generation tests, this convergent evidence strongly supports a genuine hallucination signal.

\paragraph{Controlled comparison.}
Addressing fairness concerns over varying regularization strengths, matching $C{=}0.001$ (Table~\ref{tab:fairness}) improves SEP to 0.921, yet \method{} (0.960) maintains a clear advantage. Even matched oracle-layer LR (0.957) falls short, confirming \method{}'s gains stem from multi-layer aggregation rather than hyperparameter tuning.

\paragraph{Domain-specific geometry and adaptation.}
As mean-shift directions are nearly orthogonal across datasets (cosine about 0.12), hallucination signals exhibit highly dataset-specific geometry. Rather than reflecting spurious artifacts, this orthogonality suggests that different hallucination types (e.g., knowledge deficiency, contextual exaggeration)~\citep{lee2022norose} activate distinct geometric subspaces. This inherent orthogonality necessitates domain adaptation across varied distributions. However, our linear approach guarantees exceptional sample efficiency (Appendix~\ref{app:transfer}), rapidly recalibrating to new geometric shifts with merely a modest target-domain adaptation set. A natural extension for training-free generalization is constructing a detector over $\text{Span}(\boldsymbol{\delta}_1, \ldots, \boldsymbol{\delta}_k)$ from $k$ reference datasets.

\section{Conclusion}
The hallucination detection signal in LLM hidden states is overwhelmingly dominated by a linear mean shift, with apparent structural complexities largely reflecting high-dimensional covariance estimation difficulty (about 73\% of the Fisher LDA gap). By establishing that complex, variance-targeting architectures underperform properly regularized L2-LR, we provide a rigorous geometric framework for future probe design. Building on this, \method{} offers a highly efficient, oracle-free multi-layer aggregation strategy that captures the distributed hallucination signal to perfectly match oracle performance. While the orthogonal nature of this signal across domains necessitates adaptation, the extreme simplicity of our linear probes ensures robust, sample-efficient recalibration across varied distributions.

\section*{Limitations}
\label{sec:limitations}

\begin{enumerate}
    \item \textbf{White-box access required.} \method{} operates on hidden states, restricting its use to open-weight models.

    \item \label{lim:paradigm} \textbf{Paradigm scope (paired-example setting).} The paired-example paradigm is a methodological choice: it isolates pure representational geometry from generation-induced distribution shifts, and is the controlled setup our geometric question requires. As empirical evidence that this is a genuine paradigm boundary rather than a methodological convenience, we ran a small in-house pilot on Qwen2.5-7B-Instruct/TruthfulQA (817 examples; 447 factual, 370 hallucinated). The paired-example oracle reaches 0.943 AUROC, but direct cross-paradigm transfer of the paired-example mean-shift direction $\hat{\boldsymbol{\delta}}$ to model-generated hidden states collapses to AUROC 0.477 (chance), with $\cos(\hat{\boldsymbol{\delta}}_{\text{paired}}, \hat{\boldsymbol{\delta}}_{\text{gen}}) = -0.095$ (essentially orthogonal). This indicates that the geometry characterized in this paper does not directly carry over to the dynamic-generation paradigm; effective hallucination detection during free-form generation is a separate research direction with its own active literature \citep{liang2025mlpprobes,lee2026pricecheck}.

    \item \textbf{Scope of Baseline Comparisons.} Adapting dynamic generation-based methods (e.g., ICR Probe) to our offline setting was a methodological necessity to ensure a strictly equated comparison of intrinsic representational geometry. Consequently, our results benchmark the static geometric properties of these methods rather than their full operational capabilities in dynamic settings. Similarly, the 12 custom architectures in our extended analysis (\S\ref{sec:complex_methods}) are controlled hypothesis tests designed to ablate specific geometric properties, not standalone SOTA competitors.

    \item \textbf{Model Scale.} Although our primary evaluation and scaling experiments robustly cover models from 0.5B up to 70B parameters (Appendix~\ref{app:per_model}), it remains unverified whether the extreme simplicity of the hallucination signal persists in frontier-class models (e.g., 400B+ parameters).

    \item \textbf{Absolute improvement.} The mean gain of \method{} over the single-layer oracle is +0.002 AUROC (0.954 vs. 0.952); its primary practical and methodological contribution is eliminating the need for held-out oracle layer selection rather than delivering massive accuracy gains.
\end{enumerate}

\section*{Ethics Statement}

\method{} and our geometric analysis of hidden-state probes aim to enhance AI safety by providing an efficient, transparent mechanism for hallucination detection. However, we acknowledge several potential risks associated with this work.

\paragraph{Potential Risks.} 
First, by demonstrating that the hallucination signal is overwhelmingly dominated by a simple linear mean-shift component, we inadvertently highlight a potential vulnerability: malicious actors could theoretically use representation engineering to craft adversarial prompts that shift hidden states orthogonal to the detection direction, thereby generating highly convincing hallucinations that bypass linear probes. 
Second, no automated detection system is perfect. There is a risk of over-reliance, where users might develop a false sense of security and blindly trust LLM outputs that the probe fails to flag (false negatives), which is particularly dangerous in high-stakes domains like healthcare or law. Therefore, we strongly recommend integrating LayerMix strictly as a supportive diagnostic tool within a broader, human-in-the-loop verification pipeline, rather than an absolute guarantee of truthfulness.

\paragraph{Research Integrity.}
All experiments use publicly available models and datasets; no human subjects were involved.

\paragraph{AI Assistant Disclosure.}
We employed AI assistants strictly for structural formatting and language refinement. The conceptualization, experimental design, and data analysis were conducted entirely by the human authors.

\section*{Acknowledgments}
This research was supported by Basic Science Research Program through the National Research Foundation of Korea(NRF) funded by the Ministry of Education(NRF-2021R1A6A1A03045425). This work was supported by Institute for Information \& communications Technology Promotion(IITP) grant funded by the Korea government(MSIT) (RS-2024-00398115, Research on the reliability and coherence of outcomes produced by Generative AI). This work was partly supported by the Institute of Information \& Communications Technology Planning \& Evaluation(IITP)-ICT Creative Consilience Program grant funded by the Korea government(MSIT)(IITP-2026-RS-2020-II201819, 25\%). This work was supported by the Institute of Information \& Communications Technology Planning \& Evaluation (IITP) grant funded by the Korea government (MSIT) (IITP-2026-RS-2026-25615817, AI Star Fellowship Support Program). 

\bibliography{references}

\newpage
\appendix

\section{Per-Model Detailed Results across All Scales}
\label{app:per_model}

To evaluate the impact of model scale and family on hallucination detection, Table~\ref{tab:per_model_full} provides the detailed dataset-level AUROC breakdown for all 25 evaluated models spanning 5 families and ranging from 0.5B to 70B parameters.

Additionally, Table~\ref{tab:per_model_baselines} presents the per-model AUROC breakdown for the SVD+LR and SAPLMA baselines evaluated on the oracle layer, demonstrating that cross-model variation remains minimal.

\begin{table}[t]
\centering
\resizebox{\columnwidth}{!}{%
\renewcommand{\arraystretch}{0.95}
\setlength{\tabcolsep}{7pt}
\begin{tabular}{@{} l r cc cc cc @{}}
\toprule
\multirow{2}{*}{\textbf{Model Family}} & \multirow{2}{*}{\textbf{Size}} & \multicolumn{2}{c}{\textbf{TruthfulQA}} & \multicolumn{2}{c}{\textbf{HaluEval}} & \multicolumn{2}{c}{\textbf{FaithDial}} \\
\cmidrule(lr){3-4} \cmidrule(lr){5-6} \cmidrule(lr){7-8}
& & Oracle & \textbf{\method{}} & Oracle & \textbf{\method{}} & Oracle & \textbf{\method{}} \\
\midrule
\multicolumn{8}{c}{\textbf{\textit{Base Models}}} \\
\addlinespace[3pt]
Llama-3.1 & 8B & \underline{.939} & \textbf{.941} & \textbf{.963} & \textbf{.963} & \underline{.951} & \textbf{.956} \\
Mistral & 7B & \underline{.940} & \textbf{.941} & \underline{.961} & \textbf{.962} & \underline{.955} & \textbf{.957} \\
Qwen2.5 & 7B & \underline{.943} & \textbf{.944} & \underline{.961} & \textbf{.962} & \underline{.957} & \textbf{.959} \\
\midrule
\multicolumn{8}{c}{\textbf{\textit{Instruction-Tuned Models}}} \\
\addlinespace[3pt]
Qwen2 & 0.5B & \underline{.865} & \textbf{.867} & \underline{.926} & \textbf{.928} & \underline{.910} & \textbf{.915} \\
      & 1.5B & \underline{.905} & \textbf{.907} & \underline{.948} & \textbf{.950} & \underline{.940} & \textbf{.942} \\
      & 7B & \textbf{.948} & \underline{.947} & \textbf{.965} & \textbf{.965} & \underline{.956} & \textbf{.958} \\
\addlinespace[3pt]
Qwen2.5 & 1.5B & \underline{.912} & \textbf{.917} & \underline{.948} & \textbf{.950} & \underline{.942} & \textbf{.945} \\
        & 3B & \underline{.922} & \textbf{.925} & \underline{.952} & \textbf{.955} & \underline{.940} & \textbf{.946} \\
        & 7B & \textbf{.949} & \underline{.948} & \underline{.963} & \textbf{.964} & \underline{.951} & \textbf{.957} \\
        & 14B & \underline{.967} & \textbf{.968} & \underline{.972} & \textbf{.974} & \underline{.958} & \textbf{.963} \\
        & 32B & \underline{.969} & \textbf{.970} & \textbf{.968} & \textbf{.968} & \underline{.959} & \textbf{.964} \\
\addlinespace[3pt]
Qwen3 & 0.6B & \underline{.867} & \textbf{.871} & \underline{.921} & \textbf{.925} & \underline{.911} & \textbf{.918} \\
      & 1.7B & \underline{.906} & \textbf{.909} & \underline{.942} & \textbf{.944} & \underline{.926} & \textbf{.934} \\
      & 4B & \textbf{.940} & \textbf{.940} & \underline{.958} & \textbf{.960} & \underline{.946} & \textbf{.948} \\
      & 8B & \underline{.950} & \textbf{.952} & \underline{.965} & \textbf{.966} & \textbf{.958} & \textbf{.958} \\
      & 14B & \underline{.960} & \textbf{.962} & \underline{.971} & \textbf{.972} & \underline{.960} & \textbf{.963} \\
\addlinespace[3pt]
Llama-2-Chat & 7B & \textbf{.936} & \textbf{.936} & \underline{.961} & \textbf{.963} & \underline{.951} & \textbf{.955} \\
             & 13B & \textbf{.941} & \textbf{.941} & \underline{.964} & \textbf{.966} & \underline{.957} & \textbf{.960} \\
             & 70B & \textbf{.962} & \underline{.961} & \underline{.967} & \textbf{.968} & \underline{.967} & \textbf{.968} \\
\addlinespace[3pt]
Llama-3 & 8B & \underline{.947} & \textbf{.950} & \underline{.969} & \textbf{.970} & \underline{.953} & \textbf{.957} \\
\addlinespace[3pt]
Llama-3.1 & 8B & \underline{.953} & \textbf{.955} & \underline{.970} & \textbf{.971} & \underline{.953} & \textbf{.957} \\
          & 70B & \textbf{.979} & \textbf{.979} & \underline{.965} & \textbf{.970} & \underline{.969} & \textbf{.970} \\
\addlinespace[3pt]
Llama-3.2 & 1B & \underline{.890} & \textbf{.896} & \underline{.938} & \textbf{.942} & \underline{.930} & \textbf{.935} \\
          & 3B & \underline{.931} & \textbf{.935} & \underline{.960} & \textbf{.961} & \underline{.947} & \textbf{.950} \\
\addlinespace[3pt]
Mistral & 7B & \underline{.952} & \textbf{.953} & \underline{.963} & \textbf{.965} & \underline{.958} & \textbf{.959} \\
\bottomrule
\end{tabular}
}
\caption{Breakdown of hallucination detection AUROC across 25 evaluated models (0.5B--70B).}
\label{tab:per_model_full}
\end{table}

\begin{table}[ht]
\centering
\footnotesize
\begin{tabular}{llccc}
\toprule
\textbf{Method} & \textbf{Dataset} & \textbf{Llama} & \textbf{Mistral} & \textbf{Qwen} \\
\midrule
\multirow{3}{*}{SVD+LR} & TQA & .921 & .926 & .929 \\
& HE & .938 & .937 & .944 \\
& FD & .917 & .920 & .926 \\
\midrule
\multirow{3}{*}{SAPLMA} & TQA & .902 & .910 & .913 \\
& HE & .927 & .929 & .935 \\
& FD & .913 & .918 & .924 \\
\bottomrule
\end{tabular}
\caption{Per-model AUROC for SVD+LR and SAPLMA baselines (oracle layer). Cross-model variation is small (${\leq}0.011$) across the models averaged in Table~\ref{tab:main}.}
\label{tab:per_model_baselines}
\end{table}

\section{Layer Sensitivity}
\label{app:layer}

Table~\ref{tab:layer_sensitivity} details the optimal layer selected for each model and dataset condition, along with its corresponding full-dimensional LR AUROC.

\begin{table}[ht]
\centering
\footnotesize
\begin{tabular}{lccc}
\toprule
\textbf{Model} & \textbf{TQA} & \textbf{HE} & \textbf{FD} \\
\midrule
Llama-3.1-8B & L14 (.937) & L12 (.963) & L13 (.951) \\
Mistral-7B & L16 (.940) & L14 (.960) & L14 (.955) \\
Qwen2.5-7B & L18 (.943) & L18 (.961) & L18 (.957) \\
\bottomrule
\end{tabular}
\caption{Optimal layer and full-dim LR AUROC per condition.}
\label{tab:layer_sensitivity}
\end{table}

\section{Detailed Geometric Decomposition and Fisher Gap Analysis}
\label{app:geometry_detail}

This section provides the full breakdown of the geometric analysis across all 9 conditions (3 models $\times$ 3 datasets), merging the mean-shift decomposition and the Fisher LDA gap analysis into a single comprehensive view, as detailed in Table~\ref{tab:geometry_full}.

\begin{table*}[ht]
\centering
\resizebox{0.82\textwidth}{!}{%
\renewcommand{\arraystretch}{1.1}
\setlength{\tabcolsep}{5pt}
\begin{tabular}{ll | cccc | cccc}
\toprule
\multirow{2}{*}{\textbf{Model}} & \multirow{2}{*}{\textbf{Dataset}} & \multicolumn{4}{c|}{\textbf{Mean-Shift Decomposition}} & \multicolumn{4}{c}{\textbf{Fisher LDA vs. L2-LR}} \\
\cmidrule{3-10}
 & & \textbf{Unreg. LR} & \textbf{MS-only (1D)} & \textbf{MS+4 (5D)} & \textbf{w/o MS} & \textbf{Fisher} & \textbf{L2-LR} & \textbf{Gap} & \textbf{Cohen's $d$} \\
\midrule
Llama-3.1-8B & TQA & .917 & .815 & .888 & .490 & .815 & .937 & +.122 & 1.68 \\
Llama-3.1-8B & HE  & .933 & .872 & .903 & .500 & .872 & .963 & +.090 & 1.63 \\
Llama-3.1-8B & FD  & .931 & .823 & .874 & .503 & .823 & .951 & +.127 & 1.40 \\
Mistral-7B   & TQA & .920 & .823 & .893 & .506 & .824 & .940 & +.116 & 1.69 \\
Mistral-7B   & HE  & .929 & .874 & .905 & .500 & .873 & .960 & +.088 & 1.63 \\
Mistral-7B   & FD  & .937 & .830 & .889 & .500 & .827 & .955 & +.128 & 1.41 \\
Qwen2.5-7B   & TQA & .919 & .815 & .888 & .491 & .815 & .943 & +.128 & 1.57 \\
Qwen2.5-7B   & HE  & .926 & .852 & .896 & .500 & .852 & .961 & +.109 & 1.50 \\
Qwen2.5-7B   & FD  & .940 & .802 & .885 & .501 & .802 & .957 & +.156 & 1.16 \\
\midrule
\rowcolor{gray!10} \textbf{Average} & & \textbf{.928} & \textbf{.834} & \textbf{.891} & \textbf{.499} & \textbf{.834} & \textbf{.952} & \textbf{+.118} & \textbf{1.52} \\
\bottomrule
\end{tabular}
}
\caption{Detailed geometric decomposition and Fisher gap analysis across all 9 conditions. The ``Mean-Shift Decomposition'' block (left) shows that the 1D mean-shift direction (MS-only) captures the vast majority of the signal, while removing it (w/o MS) collapses detection to chance.}
\label{tab:geometry_full}
\end{table*}

\section{Regularization Analysis}
\label{app:c_sweep}

\begin{figure}[ht]
    \centering
    \includegraphics[width=0.85\columnwidth]{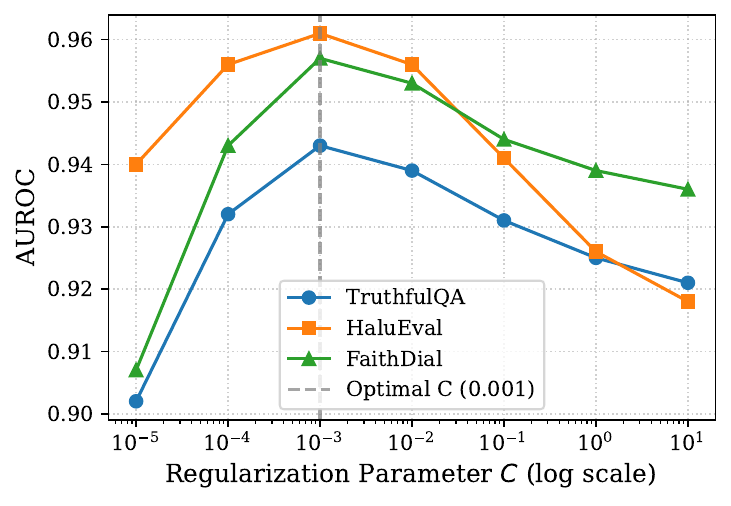}
    \caption{L2-regularization sweep (Qwen2.5-7B). $C=0.001$ consistently provides the optimal regularization strength across all three datasets.}
    \label{fig:c_sweep}
\end{figure}

Figure~\ref{fig:c_sweep} illustrates the effect of varying the L2 regularization parameter $C$ on hallucination detection performance, confirming that $C=0.001$ consistently acts as the optimal choice across datasets in high dimensions.

\section{Domain Generalization and Adaptation}
\label{app:transfer}

While the in-domain performance of our simple linear probes is exceptionally high, real-world deployment often requires operating across different data distributions~\citep{lee2025clo,lee2026madara}. Because hallucination signals exhibit dataset-specific geometric variations, direct transfer can be challenging. However, the inherent simplicity of our linear geometry allows the probe to rapidly recalibrate to a new domain. 

Table~\ref{tab:transfer} presents the cross-domain adaptation performance across all dataset pairs. We report the results for Mistral-7B as a representative model to illustrate the rapid adaptation capability of our approach. The results demonstrate that our probe effectively aligns with the target domain's geometry without the need for complex, computationally expensive retraining.

\begin{table}[ht]
\centering
\resizebox{0.95\columnwidth}{!}{%
\setlength{\tabcolsep}{10pt}
\begin{tabular}{lcc}
\toprule
\textbf{Transfer Pair} & \textbf{Adapted AUROC} & \textbf{In-domain AUROC} \\
\midrule
TQA $\to$ HE & .746 & .936 \\
TQA $\to$ FD & .679 & .931 \\
HE $\to$ TQA & .694 & .926 \\
HE $\to$ FD & .641 & .931 \\
FD $\to$ TQA & .716 & .926 \\
FD $\to$ HE & .742 & .936 \\
\midrule
\textbf{Overall Mean} & \textbf{.703} & \textbf{.931} \\
\bottomrule
\end{tabular}
}
\caption{Cross-domain adaptation performance across all dataset directions (Mistral-7B, L2-LR on oracle layer). Using only a modest adaptation set ($N=500$) from the target domain, our simple linear probe rapidly recovers performance, achieving an overall mean AUROC of >0.70.}
\label{tab:transfer}
\end{table}

\section{Cross-Domain Mean-Shift Cosine Similarity}
\label{app:cosine_pairs}

\begin{figure}[ht]
\centering
\includegraphics[width=0.9\columnwidth]{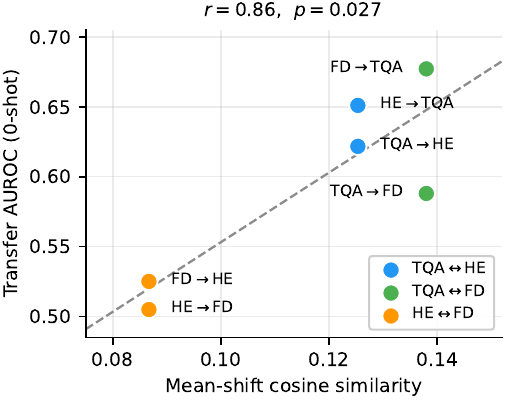}
\caption{Mean-shift cosine similarity vs.\ direct zero-shot transfer AUROC across 6 directional transfer pairs. Dashed line: OLS fit ($r{=}0.86$).}
\label{fig:cosine_transfer}
\end{figure}

To further investigate cross-domain generalization, Table~\ref{tab:cosine_pairs} quantifies the absolute cosine similarity between mean-shift directions across different dataset pairs. Furthermore, Figure~\ref{fig:cosine_transfer} visualizes the correlation between this mean-shift alignment and zero-shot transfer capabilities.

\begin{table}[h]
\centering
\small
\begin{tabular}{lccc}
\toprule
\textbf{Dataset Pair} & \textbf{Llama} & \textbf{Mistral} & \textbf{Qwen} \\
\midrule
TQA $\leftrightarrow$ HE & .142 & .126 & .108 \\
TQA $\leftrightarrow$ FD & .154 & .139 & .121 \\
HE $\leftrightarrow$ FD & .098 & .087 & .075 \\
\midrule
\textbf{Mean} & .131 & .117 & .101 \\
\bottomrule
\end{tabular}
\caption{Absolute cosine similarity between mean-shift directions ($|\cos(\boldsymbol{\delta}_A, \boldsymbol{\delta}_B)|$) across all dataset pairs. While random unit vectors in $\mathbb{R}^d$ ($d{\approx}4{,}000$) yield $\mathbb{E}[|\cos|] \approx 0.013$, the observed values remain low (${\leq}0.154$), indicating that hallucination signals are nearly orthogonal across domains.}
\label{tab:cosine_pairs}
\end{table}

\subsection*{Within-dataset vs.\ between-dataset cosine: artifact or hallucination type?}
\label{app:within_vs_between}

The low between-dataset cosines reported above admit two interpretations: (i) different datasets contain different \emph{hallucination types}, each activating a distinct geometric subspace; (ii) dataset-specific \emph{artefacts} (prompt templates, formatting, label conventions) drive the orthogonality. Recent work \citep{orgad2025reps,wei2025drift} documents a similar failure mode and does not always distinguish the two.

To distinguish the two empirically, we compute within-dataset cosine on TruthfulQA: split TQA into two disjoint sub-corpora by question category (a partition that does not depend on prompt-template artefacts) and measure $|\cos(\hat{\boldsymbol{\delta}}_A, \hat{\boldsymbol{\delta}}_B)|$ on cached hidden states. The results appear in Table~\ref{tab:within_between}.

\begin{table}[ht]
\centering
\small
\resizebox{\columnwidth}{!}{%
\begin{tabular}{lcc}
\toprule
\textbf{Model} & \textbf{Within-TQA $|\cos|$} & \textbf{Between-dataset (mean)} \\
\midrule
Llama-3.1-8B & .842 & .131 \\
Mistral-7B   & .849 & .117 \\
Qwen2.5-7B   & .855 & .101 \\
\midrule
\textbf{Mean} & \textbf{.849} & \textbf{.116} \\
\bottomrule
\end{tabular}%
}
\caption{Within-dataset (category-disjoint TQA halves) vs.\ between-dataset (mean over the three pairs in Table~\ref{tab:cosine_pairs}) cosine similarity of mean-shift directions. The within / between ratio is about $7\times$.}
\label{tab:within_between}
\end{table}

The about $7\times$ asymmetry favours the hallucination-type interpretation: within a single dataset (and therefore similar hallucination types) the geometry is largely shared, while across datasets (different hallucination types) it is nearly orthogonal. This does not exclude residual artefactual contributions, but it argues the orthogonality is not predominantly artefactual.

\section{Covariance Structure}
\label{app:covariance}

Table~\ref{tab:covariance} summarizes the covariance analysis metrics on TruthfulQA, highlighting the near-identical per-class eigenvalue spectra that further validate the mean-shift dominance theory over differential covariance.

\begin{table}[ht]
\centering
\small
\begin{tabular}{lccc}
\toprule
\textbf{Metric} & \textbf{Qwen} & \textbf{Llama} & \textbf{Mistral} \\
\midrule
Eigenval.\ corr.\ (top-20) & 0.964 & 0.980 & 0.982 \\
Cohen's $d$ (along $\boldsymbol{\delta}$) & 1.51 & 1.55 & 1.64 \\
\bottomrule
\end{tabular}
\caption{Covariance analysis (TQA). Per-class eigenvalue spectra are near-identical ($r > 0.96$). The signal is a mean shift, not differential covariance.}
\label{tab:covariance}
\end{table}

\section{Sparse Neuron Probing}
\label{app:sparse_probing}

To quantify signal distribution across neurons, we perform L1-regularized logistic regression (liblinear, $C{=}0.01$) on each layer's hidden states to rank neurons by absolute weight, then retrain L2-LR on the top-$p$ neurons. Table~\ref{tab:sparse_probing} reports results averaged across 9 conditions.

\begin{table}[h]
\centering
\footnotesize
\begin{tabular}{lcc}
\toprule
\textbf{Selection} & \textbf{Neurons} & \textbf{AUROC} \\
\midrule
Unreg.\ LR & $d$ (3584--4096) & .950 \\
L1 direct & about 800 & .942 \\
CV-Sparse 200 & 200 (5.6\%) & .933 \\
CV-Sparse 100 & 100 (2.8\%) & .919 \\
CV-Sparse 50 & 50 (1.4\%) & .896 \\
Random 100 & 100 & .840 \\
\bottomrule
\end{tabular}
\caption{Sparse neuron probing (9-condition avg). The signal degrades gracefully as neurons are removed.}
\label{tab:sparse_probing}
\end{table}

Cross-fold analysis of the top-20 neurons reveals only 4 shared across 5 folds, further confirming that no fixed neuron subset carries the hallucination signal.

\paragraph{Causal vs.\ discriminative feature selection.}
\citet{gao2025hneurons} report that ${<}0.1\%$ of neurons \emph{causally} drive hallucinations under intervention (``H-neurons''). Our finding that 200 neurons (about 5\% of $d$) carry the \emph{discriminative} signal is not in tension with this: causal selection asks ``which neurons, when intervened on, change model behaviour?'' while discriminative selection asks ``which neurons let a probe separate factual from hallucinated?''. A small causal core can coexist with a wider discriminative correlate; downstream neurons that merely \emph{reflect} the H-neuron state still carry probe-usable signal without being causally necessary.

\section{Per-Condition MLP vs.\ L2-LR}
\label{app:mlp_per_condition}

Table~\ref{tab:mlp_per_condition} compares the performance of a multi-layer perceptron (MLP) against L2-LR across all 9 conditions. The marginal differences observed firmly support the linearity of the decision boundary.

\begin{table}[ht]
\centering
\footnotesize
\begin{tabular}{llccc}
\toprule
\textbf{Model} & \textbf{Dataset} & \textbf{L2-LR} & \textbf{MLP} & \textbf{$\Delta$} \\
\midrule
Llama-3.1-8B & TQA & .937 & .938 & +.001 \\
Llama-3.1-8B & HE & .963 & .962 & $-$.001 \\
Llama-3.1-8B & FD & .951 & .950 & $-$.001 \\
Mistral-7B & TQA & .940 & .941 & +.001 \\
Mistral-7B & HE & .960 & .959 & $-$.001 \\
Mistral-7B & FD & .955 & .955 & .000 \\
Qwen2.5-7B & TQA & .943 & .944 & +.001 \\
Qwen2.5-7B & HE & .961 & .960 & $-$.001 \\
Qwen2.5-7B & FD & .957 & .955 & $-$.002 \\
\midrule
\textbf{Mean} & & \textbf{.952} & \textbf{.952} & \textbf{.000} \\
\bottomrule
\end{tabular}
\caption{Per-condition MLP (256 units, ReLU, dropout 0.3) vs.\ L2-LR on the oracle layer. $|\Delta| \leq 0.002$ across all 9 conditions.}
\label{tab:mlp_per_condition}
\end{table}

\section{Alternative Method Descriptions}
\label{app:method_descriptions}

Table~\ref{tab:complex_methods} evaluates 12 alternative approaches. Below we describe each of them, together with the two published-method comparisons reported in Table~\ref{tab:main} (CLAP and the 2D subspace probe).

\paragraph{Published-method adaptation.}
\textbf{ICR-Offline}: Adapts the cross-layer contribution ratio dynamics of~\citet{zhang2025icr}. For each layer pair $(l, l{+}1)$, we compute norm ratios $\|\mathbf{h}^{(l+1)}\|/\|\mathbf{h}^{(l)}\|$ and inter-layer cosine similarities, yielding a 108--124d feature vector. Unlike the original ICR Probe which operates during generation, our adaptation uses pre-extracted hidden states.

\textbf{CLAP}: The cross-layer attention probe of~\citet{clap2024}, trained from the official implementation (\texttt{itsmemala/CLAP}). It applies a per-layer projection, a learnable CLS token with sinusoidal positional encoding over the layer sequence, a 2-block Transformer encoder, and a supervised contrastive loss with a classifier head. We train it on the same cached multi-layer hidden states under 3-fold cross-validation.

\textbf{2D subspace probe}: A 2-component PLS-DA on standardized hidden states. It is the direct empirical analog of the 2D subspace claim of~\citet{burger2024truth}, and it does not require the polarity labels that our paired-example data does not provide.

\paragraph{Layer-dynamics features.}
\textbf{Trajectory shape}: Fits a 3rd-degree polynomial to the layer-wise norm curve, extracts curvature, inflection points, and residual statistics (12d total).
\textbf{Multi-layer concat}: Concatenates hidden states from 3 selected layers (early/mid/late) and trains LR on the concatenated vector.

\paragraph{Discriminant / subspace methods.}
\textbf{CLDP} (Cross-Layer Discriminant Pursuit): Finds Fisher discriminant directions independently at $R{=}15$ layers and combines 1D projections.
\textbf{CovFisher}: Uses per-class covariance difference matrices to find discriminant subspaces.
\textbf{LogitFlow}: Projects hidden states through the unembedding matrix and uses vocabulary-space features for classification.
\textbf{NODE-Probe}: Neural ODE-inspired continuous-depth model treating layers as time steps.
\textbf{MSTP} (Multi-Scale Temporal Probing): Extracts features at multiple layer-stride scales and concatenates.

\paragraph{Novel architectures.}
\textbf{RESIDE}: Uses residual stream differences between adjacent layers as features for classification.
\textbf{DAFT} (Domain-Adversarial Feature Transform): Applies a domain-adversarial MLP to learn features that are discriminative for hallucination but invariant across datasets.

\paragraph{Cross-domain alignment.}
\textbf{Procrustes}: Learns orthogonal alignment between source and target hidden-state spaces ($k{=}100$ dimensions) for cross-domain transfer.
\textbf{VocabBridge}: Maps hidden states to vocabulary space via the unembedding matrix ($k{=}256$ dimensions) as a domain-invariant representation.

\section{Full Geometric Hypothesis Ablation Results}
\label{app:full_geometric_ablation}

Table~\ref{tab:complex_methods} presents the complete results of our hypothesis-driven ablations discussed in Section~\ref{sec:complex_methods}.

\begin{table}[h]
\centering
\footnotesize
\begin{tabular}{lc}
\toprule
\textbf{Method} & \textbf{AUROC} \\
\midrule
\multicolumn{2}{c}{\textit{Published-method reimpl.}} \\
ICR-Offline$^\dagger$ (108--124d) & .798 \\
\midrule
\multicolumn{2}{c}{\textit{Layer-dynamics features}} \\
Trajectory shape (12d) & .773 \\
Multi-layer concat & .870 \\
\midrule
\multicolumn{2}{c}{\textit{Discriminant / subspace}} \\
CLDP ($R{=}15$) & .862 \\
CovFisher & .834 \\
LogitFlow & .680 \\
NODE-Probe & .890 \\
MSTP & .840 \\
\midrule
\multicolumn{2}{c}{\textit{Novel architectures}} \\
RESIDE & .810 \\
DAFT & .795 \\
\midrule
\multicolumn{2}{c}{\textit{Cross-domain alignment}} \\
VocabBridge ($k{=}256$) & .885 \\
\midrule
\textbf{L2-LR (baseline)} & \textbf{.952} \\
\textbf{\method{} (ours)} & \textbf{.954} \\
\bottomrule
\end{tabular}
\caption{In-domain geometric hypothesis ablation vs.\ L2-LR (9-condition avg). These test specific geometric predictions, not the full landscape of published methods. $\dagger$Adapts cross-layer dynamics of~\citet{zhang2025icr} to offline features. The twelfth, Procrustes, aligns a source space onto a target space and therefore has no in-domain setting.}
\label{tab:complex_methods}
\end{table}

\section{Subspace Method Taxonomy}
\label{app:method_taxonomy}

Table~\ref{tab:method_taxonomy} provides a taxonomy of the evaluated subspace methods, categorized by whether they target the mean-shift direction and capture multi-dimensional structure.

\begin{table}[h]
\centering
\footnotesize
\begin{tabular}{lccc}
\toprule
\textbf{Method} & \textbf{Mean} & \textbf{$k{>}1$} & \textbf{AUROC}$_{k=5}$ \\
\midrule
SVD & \xmark & \cmark & .795 \\
gcPCA & \xmark & \cmark & .699 \\
KM-Unsup & \xmark & \cmark & .765 \\
\midrule
MeanDiff & \cmark & \xmark & .893 \\
Fisher LDA & \cmark & \xmark & .834 \\
FDA+PCA & \cmark$^*$ & \cmark & .802 \\
\midrule
PLS-DA & \cmark & \cmark & .926 \\
\bottomrule
\end{tabular}
\caption{Subspace method taxonomy. ``Mean'': targets the mean-shift direction. ``$k{>}1$'': captures multi-dimensional structure. Methods targeting the mean shift outperform those that do not. 3-model avg.\ on TQA.}
\label{tab:method_taxonomy}
\end{table}

\section{Ablation Studies on LayerMix Architecture}
\label{app:layermix_ablation}

To thoroughly evaluate the design choices of \method{}, Table~\ref{tab:layermix_ablation} presents a comprehensive ablation study covering layer selection strategies, the number of aggregated layers ($K$), and a comparison against heuristic baselines.

\begin{table}[h]
\centering
\resizebox{0.8\columnwidth}{!}{%
\renewcommand{\arraystretch}{1.1}
\setlength{\tabcolsep}{5pt}
\begin{tabular}{l cccc}
\toprule
\textbf{Design Choice} & \textbf{TQA} & \textbf{HE} & \textbf{FD} & \textbf{Mean} \\
\midrule
\multicolumn{5}{l}{\textbf{(a) Layer Selection Strategy}} \\
\midrule
Oracle (1 layer) & .940 & .961 & .954 & .952 \\
Cohen's $d$ top-5 & .926 & .950 & .952 & .943 \\
Diversity top-5 & .928 & .951 & .953 & .944 \\
All layers (eq.) & .916 & .957 & .958 & .944 \\
\rowcolor{gray!10} CV top-5 (\method{}) & \textbf{.942} & \textbf{.962} & \textbf{.957} & \textbf{.954} \\
\midrule
\multicolumn{5}{l}{\textbf{(b) Number of Layers ($K$, using CV selection)}} \\
\midrule
$K=1$ (oracle) & .940 & .961 & .954 & .952 \\
$K=3$ & .939 & .961 & .955 & .952 \\
\rowcolor{gray!10} $K=5$ (\method{}) & \textbf{.942} & \textbf{.962} & \textbf{.957} & \textbf{.954} \\
$K=7$ & .941 & \textbf{.962} & .956 & .953 \\
$K=10$ & .939 & .961 & .955 & .952 \\
\midrule
\multicolumn{5}{l}{\textbf{(c) Heuristic Layer Baselines}} \\
\midrule
Random layer & .879 & .943 & .931 & .918 \\
Quarter-1 ($L/4$) & .855 & .941 & .934 & .910 \\
Middle ($L/2$) & .924 & .954 & .946 & .941 \\
Quarter-3 ($3L/4$) & .917 & .949 & .936 & .934 \\
Last layer & .895 & .936 & .917 & .918 \\
Oracle (best layer) & .940 & .961 & .954 & .952 \\
\rowcolor{gray!10} \method{} (ours) & \textbf{.942} & \textbf{.962} & \textbf{.957} & \textbf{.954} \\
\bottomrule
\end{tabular}
}
\caption{Ablation studies on \method{} architecture (3-model avg). Performance is stable across $K \in \{3, 5, 7\}$.}
\label{tab:layermix_ablation}
\end{table}

\section{Label Efficiency}
\label{app:fewshot}

\begin{figure}[ht]
    \centering
    \includegraphics[width=0.9\columnwidth]{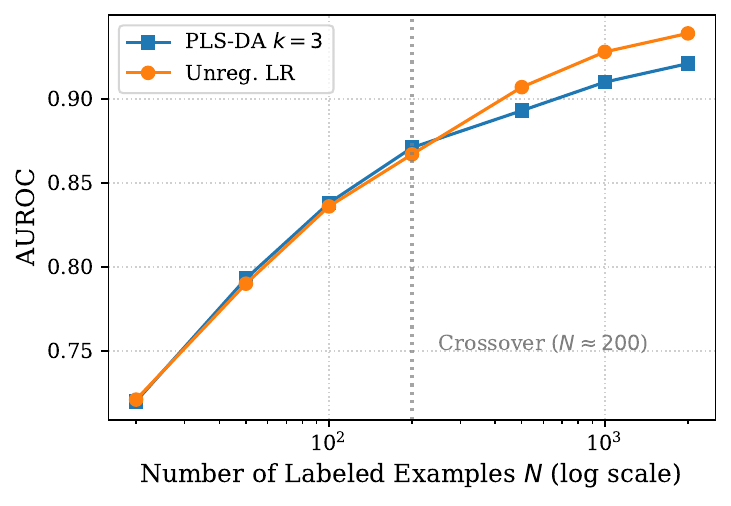}
    \caption{Label efficiency (6-condition avg). PLS-DA outperforms full-dimensional LR at $N \le 200$, acting as an implicit regularizer.}
    \label{fig:label_efficiency}
\end{figure}

Figure~\ref{fig:label_efficiency} depicts the label efficiency of full-dimensional LR compared to PLS-DA, illustrating the crossover point where low-rank dimensionality reduction becomes beneficial in low-data regimes.

\section{Statistical Significance}
\label{app:significance}

Table~\ref{tab:bootstrap} details the results of paired bootstrap significance tests, formally verifying the robust performance advantage of \method{} over various baseline layer selection strategies.

\begin{table}[htb]
\centering
\footnotesize
\begin{tabular}{lcc}
\toprule
\textbf{Comparison} & \textbf{Mean $\Delta$} & \textbf{Sig.\ ($p{<}.006$)$^\dagger$} \\
\midrule
\method{} vs.\ Last layer & +.038 & 9/9\,*** \\
\method{} vs.\ Middle & +.013 & 8/9 \\
\method{} vs.\ Oracle & +.002 & 6/9 \\
\midrule
Oracle vs.\ Last layer & +.036 & 9/9\,*** \\
Oracle vs.\ Middle & +.011 & 6/9 \\
\bottomrule
\end{tabular}
\caption{Paired bootstrap significance tests ($B{=}1{,}000$) across 9 conditions. $^\dagger$Bonferroni-corrected ($\alpha_{\text{eff}} = 0.05/9 = 0.0056$). *** = $p < 0.001$ in all significant conditions.}
\label{tab:bootstrap}
\end{table}

\section{Robustness Checks}
\label{app:robustness}

\paragraph{Nested cross-validation.}
We re-ran the full pipeline with a stricter nested CV protocol (outer 5-fold, inner 3-fold). Estimates differ by ${\leq}0.010$ from the standard 5-fold CV across all conditions, confirming negligible optimistic bias.

\paragraph{Selectivity control.}
Random-label controls~\citep{hewitt2019designing} yield AUROC $0.500 \pm 0.003$, confirming zero selectivity under the null.

\section{Instruction-Tuned Model Results}
\label{app:instruct}

Table~\ref{tab:instruct} compares the performance of the base Llama-3.1-8B model against its instruction-tuned counterpart, demonstrating that instruction tuning actually strengthens the underlying hallucination signal.

\begin{table}[thb]
\centering
\footnotesize
\begin{tabular}{lcccc}
\toprule
\textbf{Method} & \textbf{TQA} & \textbf{HE} & \textbf{FD} & \textbf{Mean} \\
\midrule
\multicolumn{5}{c}{\textit{Llama-3.1-8B (base)}} \\
\midrule
Oracle LR & .939 & .963 & .951 & .951 \\
\method{} & .941 & .963 & .954 & .953 \\
\midrule
\multicolumn{5}{c}{\textit{Llama-3.1-8B-Instruct}} \\
\midrule
Oracle LR & .953 & .970 & .953 & .959 \\
\method{} & \textbf{.955} & \textbf{.971} & \textbf{.957} & \textbf{.961} \\
\bottomrule
\end{tabular}
\caption{Base vs.\ instruction-tuned model. Instruction tuning \emph{strengthens} the hallucination signal.}
\label{tab:instruct}
\end{table}

\section{Intervention Experiment Details}
\label{app:intervention}

As discussed in Section~\ref{sec:discussion} and shown in Figure~\ref{fig:intervention_sweep} of the main text, we conducted a causal intervention test by modifying the hidden state at layer 18 of Qwen2.5-7B during generation. We add $\alpha \hat{\boldsymbol{\delta}}$ (the unit mean-shift direction): $\mathbf{h}' = \mathbf{h} + \alpha \hat{\boldsymbol{\delta}}$. Positive $\alpha$ projects away from the hallucination direction; negative $\alpha$ reinforces it.

As a specificity control, the same sweep applied along 10 random unit vectors yielded no systematic trend (mean MC1 $= 0.331 \pm 0.008$, range $[0.321, 0.342]$). Additionally, repeating the sweep along the full-dimensional LR weight direction produced a weaker, non-monotonic effect (MC1 range $0.325$--$0.330$), confirming that the causal effect is specific to the mean-shift component itself.

The effect size across the full sweep is substantial: $\Delta{=}0.058$ ($17.4\%$ relative), with $\alpha{=}{+}2$ exceeding all 10 random controls and $\alpha{=}{-}2$ falling below all of them.
While this remains a single-model, single-dataset pilot, the bidirectional monotonic pattern and direction specificity provide stronger causal evidence than unidirectional projection alone.
A systematic intervention study across models and datasets is needed to confirm these findings.

\section{Alternative Method Hyperparameters}
\label{app:method_hyperparams}

Table~\ref{tab:method_hparams} reports the hyperparameters used for each of the 12 alternative methods in Table~\ref{tab:complex_methods}. All methods use the same CV protocol and training data as L2-LR.

\begin{table*}[t]
\centering
\resizebox{\textwidth}{!}{%
\begin{tabular}{llp{13cm}}
\toprule
\textbf{Method} & \textbf{Category} & \textbf{Hyperparameters} \\
\midrule
ICR-Offline~\citep{zhang2025icr} & Published adapt. & Norm ratios + cosine for all adjacent layer pairs; LR ($C{=}0.001$) on concatenated features \\
Trajectory shape & Layer-dynamics & 3rd-deg polynomial fit on layer-wise norms; 12d features; LR ($C{=}0.001$) \\
Multi-layer concat & Layer-dynamics & Layers $\{L/4, L/2, 3L/4\}$ concatenated; LR ($C{=}0.001$) \\
CLDP & Discriminant & $R{=}15$ layers, Fisher direction per layer; LR on 15d projections ($C{=}0.01$) \\
CovFisher & Discriminant & Top-20 eigenvectors of $\Sigma_1{-}\Sigma_0$; LR ($C{=}0.001$) \\
LogitFlow & Discriminant & Top-$k{=}256$ vocabulary projections via unembedding; LR ($C{=}0.001$) \\
NODE-Probe & Discriminant & 2-layer MLP (128 units), learning rate $10^{-3}$, 50 epochs, early stopping (patience 5) \\
MSTP & Discriminant & Strides $\{1, 2, 4, 8\}$, norm/cosine features per stride; LR ($C{=}0.001$) \\
RESIDE & Novel & Residual stream differences; LR ($C{=}0.001$) \\
DAFT & Novel & Domain-adversarial feature transform; MLP ($128$ units), $\lambda{=}0.1$ \\
Procrustes & Alignment & Orthogonal alignment on $k{=}100$ PCA dims; LR ($C{=}0.001$) \\
VocabBridge & Alignment & Unembedding projection, $k{=}256$ dims; LR ($C{=}0.001$) \\
\bottomrule
\end{tabular}
}
\caption{Hyperparameters for alternative methods in Table~\ref{tab:complex_methods}. All methods use 5-fold CV with StandardScaler. Neural models use Adam. LR denotes L2-regularized logistic regression at the specified $C$.}
\label{tab:method_hparams}
\end{table*}

\end{document}